\documentclass[letterpaper, 10 pt, conference]{ieeeconf}  

\IEEEoverridecommandlockouts                              

\usepackage[utf8]{inputenc}
\usepackage{cite}
\usepackage{graphicx} 
\usepackage{balance}
\usepackage{booktabs}
\usepackage{makecell}
\usepackage{multirow}
\usepackage{url}
\usepackage{siunitx}
\title{\LARGE \bf
MSACT: Multistage Spatial Alignment \\ for Stable Low-Latency Fine Manipulation
}

\author{Xianbo Cai$^{1}$,  Hideyuki Ichiwara$^{1}$, Masaki Yoshikawa$^{1}$, and Tetsuya Ogata$^{1,2}$%
\thanks{$^{1}$Xianbo Cai, Hideyuki Ichiwara, Masaki Yoshikawa, and Tetsuya Ogata are with the Department of Intermedia Art and Science, Waseda University, Tokyo, Japan {\tt\footnotesize hourensou369@fuji.waseda.jp}}%
\thanks{$^{2}$Tetsuya Ogata is with the National Institute of Advanced Industrial Science and Technology (AIST), Tokyo, Japan {\tt\footnotesize ogata.waseda.jp}}}

\begin{document}

\maketitle
\thispagestyle{empty}
\pagestyle{empty}

\begin{abstract}
Real-world fine manipulation, particularly in bimanual manipulation, typically requires low-latency control and stable visual localization, while collecting large-scale data is costly and limited demonstrations may lead to localization drift. Existing approaches make different trade-offs: action-chunking policies such as ACT enable low-latency execution and data efficiency but rely on dense visual features without explicit spatial consistency, generative methods such as Diffusion Policy improve expressiveness but can incur iterative sampling latency, vision–language–action and voxel-based methods enhance generalization and geometric grounding but require higher computational cost and system complexity.

We introduce a multistage spatial attention module that extracts stable 2D attention points and jointly predicts future attention sequences with a temporal alignment loss. Built upon ACT with a pretrained ResNet visual prior, a multistage attention module extracts task-relevant 2D attention points as a local spatial modality for action prediction. To maintain consistent object tracking, we introduce a self-supervised objective that aligns predicted attention sequences with visual features from future frames, suppressing drift without keypoint annotations and improving stability of the vision-to-action mapping under limited data. Experiments on simulated and real-world fine manipulation tasks, conducted on the ALOHA bimanual platform, evaluate task success, attention drift, inference latency, and robustness to visual disturbances. Results indicate improvements in localization stability and task performance while maintaining low-latency inference under the tested conditions.

\end{abstract}

\section{INTRODUCTION}
Robotic fine manipulation in real-world environments, particularly in bimanual manipulation, requires policies that balance low-latency control and stable spatial localization. In deployment, perception and decision latency limits control frequency, while spatial drift during interaction can lead to contact failure and unstable execution. Meanwhile, collecting large-scale datasets is costly, making it essential for policies to maintain robust spatial alignment under limited data. Achieving low latency, data efficiency, and spatial stability simultaneously remains a central challenge in manipulation.

Existing approaches adopt different trade-offs. Imitation learning based on action chunking, such as Action Chunking with Transformers (ACT) \cite{zhao2023learning}, enables smooth low-latency execution and strong performance on real hardware. However, ACT relies on high-dimensional visual distribution without explicit spatial consistency constraints, which can lead to localization bias under visual disturbances. Diffusion Policy \cite{chi2023diffusionpolicy} models multi-modal action distributions to improve expressiveness but incur iterative sampling latency that limits real-time control frequency. Robotic vision–language–action (VLA) models, such as OpenVLA \cite{kim2024openvla}, SmolVLA \cite{shukor2025smolvla}, and $\pi$0.5 \cite{intelligence2025pi_}, leverage pretrained visual and linguistic priors for semantic generalization. However, visual tokenization and model scale impose substantial computational overhead. Some methods, such as PerAct \cite{shridhar2023perceiver}, utilize voxelized or depth-based geometric priors to enhance robustness but often depend on additional sensors, calibration, and computation, increasing system complexity. For real-world fine manipulation under limited demonstrations, there remains a need for a lightweight method that improves spatial stability while preserving low-latency control.

To address these limitations, we propose MSACT, a lightweight framework compatible with ACT, which introduces a multistage spatial alignment that incorporates explicit structured spatial representations while maintaining end-to-end inference. Instead of relying solely on dense features, a multistage spatial attention (MSA) module extracts task-relevant 2D attention points and incorporates them as a local spatial modality into the Transformer encoder, providing interpretable spatial grounding with minimal overhead. To improve temporal stability, the model jointly predicts future action chunks and future attention-point sequences. During training, attention points are re-extracted from future ground-truth frames, and a self-supervised temporal consistency loss aligns predicted and re-extracted sequences. This alignment objective reducing attention drift without requiring manual keypoint annotations, thereby stabilizing the vision-to-action mapping under limited demonstrations.

We evaluate the proposed method on simulated and real-world bimanual manipulation tasks using the ALOHA platform. Across four real-world tasks evaluated over 100 trials each, MSACT improves stage-wise and full-task completion rates compared to ACT, Diffusion Policy, SmolVLA, and $\pi0.5$. Importantly, the proposed method preserves ACT-level low-latency inference while achieving improved robustness without compromising real-time performance.

The main contributions of this work are: (1) A multistage spatial attention module that extracts local structured 2D attention points together with ResNet global visual prior for fine manipulation without requiring depth sensors or additional calibration. (2) A self-supervised temporal alignment objective that stabilizes attention sequences and suppresses drift under limited data. (3) Extensive real-world experiments on bimanual manipulation tasks, demonstrating improved spatial robustness and higher success rate while maintaining low-latency control.

\section{RELATED WORK}
Vision-based robotic fine manipulation has advanced rapidly, with approaches trading off spatial representation, generative capability, real-time performance, and data needs. Current research includes: end-to-end imitation policies, structured spatial representations, 3D geometric priors and vision–language–action (VLA) models. Despite progress, achieving accurate, robust, low-latency control with limited data and visual explainability remains challenging.

\subsection{End-to-End Imitation Learning}
End-to-end imitation learning methods directly learn control policies from image observations and have demonstrated strong performance in real-world manipulation tasks. ACT encodes multi-view RGB observations using a pretrained ResNet \cite{he2016deep}, fuses features through attention, and predicts an action chunk with a Transformer encoder–decoder. By predicting short horizons and aggregating overlapping chunks, the model achieves smooth control and reduces compounding errors while maintaining low latency. Diffusion Policy conditions a denoising network on visual features extracted by a pretrained ResNet and augmented with spatial softmax operations to retain location information. Through diffusion processes, it can model multi-modal action distributions and handle complex manipulation scenarios.

However, these approaches typically rely on high-dimensional visual features without explicit spatial representations and spatial consistency constraints, making localization failures difficult to interpret or correct. Moreover, generative policies often require multi-step iterative inference, which limits control frequency in real-time systems. In this work, we retain the low-latency advantages of ACT while modifying the visual perception mechanism. Instead of relying solely on dense high-dimensional features, we incorporate structured 2D attention point sequences as an explicit spatial modality, improving spatial robustness to environmental variations under limited data.

\subsection{Structured Spatial Representations Based Methods}
Another line of research improves visual interpretability and sample efficiency through explicit spatial representations. Methods such as Deep Spatial Autoencoders (DSAE) \cite{finn2016deep}\cite{levine2016end} and related keypoint-based visuomotor approaches \cite{zeng2018learning}\cite{zeng2021transporter} utilize spatial softmax operations to compress convolutional features into 2D coordinates, producing compact representations suitable for geometric reasoning and control. Subsequent works integrating such representations with recurrent or predictive control models demonstrate that point-based representations provide explainable interpretable spatial states and improve robustness in environmental variations and limited data\cite{ichiwara2021spatial}\cite{hiruma2022deep}\cite{cai20243d}. However, these methods typically extract attention points from the features of the final stage of vanilla 3-layer CNNs and often rely on recurrent architectures, which may lead to temporal instability in the extracted attention points and limit parallel training efficiency. These limitations motivate spatial representations that remain interpretable while supporting efficient parallel computation and stable temporal alignment.

\subsection{3D Geometric Priors and Voxel Representations}
To improve spatial consistency, many approaches incorporate explicit 3D structure derived from depth or multi-view geometry \cite{shridhar2023perceiver, liu2024voxactb}. PerAct constructs a voxel grid from calibrated RGB-D observations, converts the voxelized scene into 3D tokens, and applies a Transformer-based policy to produce actions grounded in discretized 3D space, achieving strong performance in geometrically complex manipulation tasks. Related methods \cite{chen2023polarnet}\cite{ze20243d} encode 3D point clouds using set-based networks and condition action prediction on geometry-aware features to improve invariance to appearance changes. However, such approaches typically require depth sensing and accurate calibration, and performance can degrade under noisy  (e.g., transparent or reflective objects) or missing depth measurements, increasing system complexity and deployment difficulty. In contrast, we use multi-view 2D attention points as structured spatial representations, reducing hardware and calibration requirements while maintaining spatial alignment. Temporal consistency across predicted attention sequences further couples visual alignment with action prediction to mitigate drift and improve stability.

\subsection{Robotic Foundation Models and the VLA Paradigm}
In recent years, large-scale pretrained vision-language-action (VLA) models ave emerged as a prominent direction. RT-1 \cite{brohan2022rt} and RT-2 \cite{zitkovich2023rt} train vision–language Transformer on large-scale robot datasets and employ token compression to enable real-time control from image and language instructions. OpenVLA and related open-source efforts combine powerful pretrained visual encoders with large language models, demonstrating strong cross-task generalization and highlighting the importance of adapting visual representations. More efficient VLA models such as SmolVLA and $\pi$0.5 emphasize deployability through lightweight architectures, action chunking, and system-level optimizations.

Despite their generalization capabilities, these models rely on large-scale pretraining and high-dimensional visual tokens, which can make efficient low-latency control under limited demonstrations challenging, especially for precise object-level manipulation. In this work, we compare against representative VLA baselines and investigate structured spatial alignment for fine manipulation. Instead of relying on large-scale pretraining or high-dimensional multimodal representations, our approach introduces compact attention-point sequences as an explicit spatial modality, enabling efficient and interpretable control under limited data.

\begin{figure*}
    \centering
    \includegraphics[width=1\linewidth]{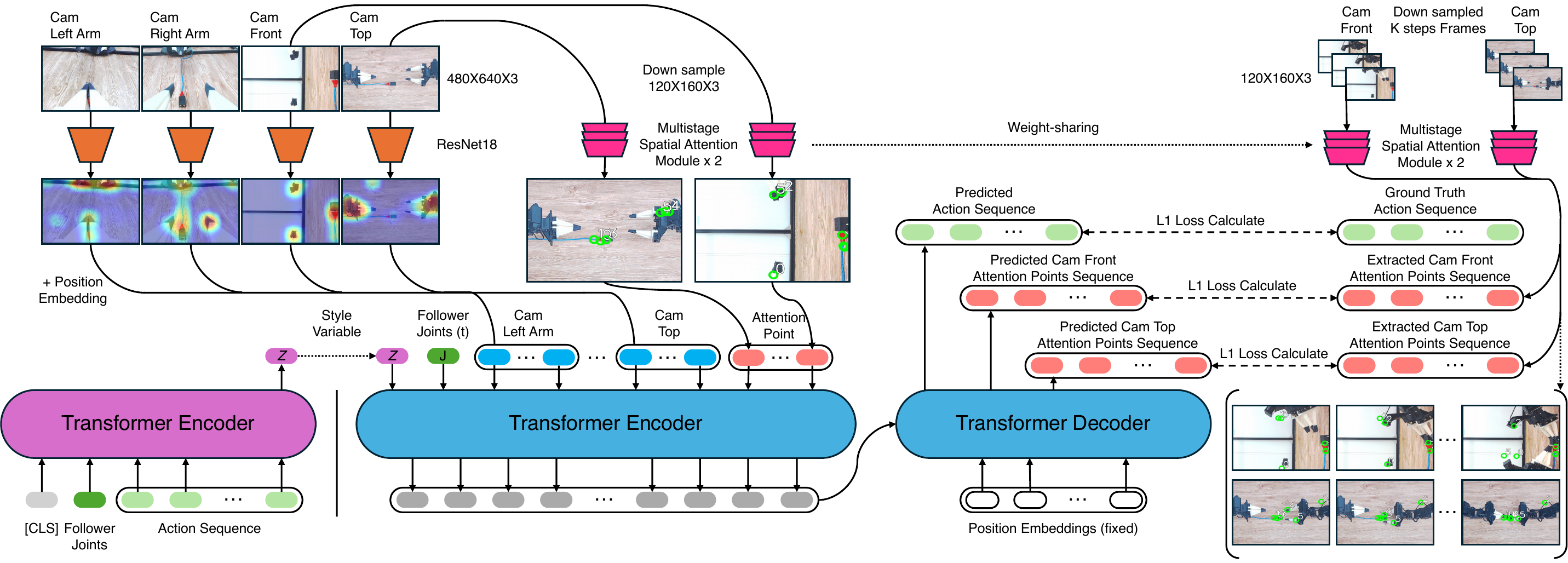}
    \caption{The overview of the proposed method.}
    \label{fig:method_overview}
\end{figure*}

\begin{figure}
    \centering
    \includegraphics[width=0.8\linewidth]{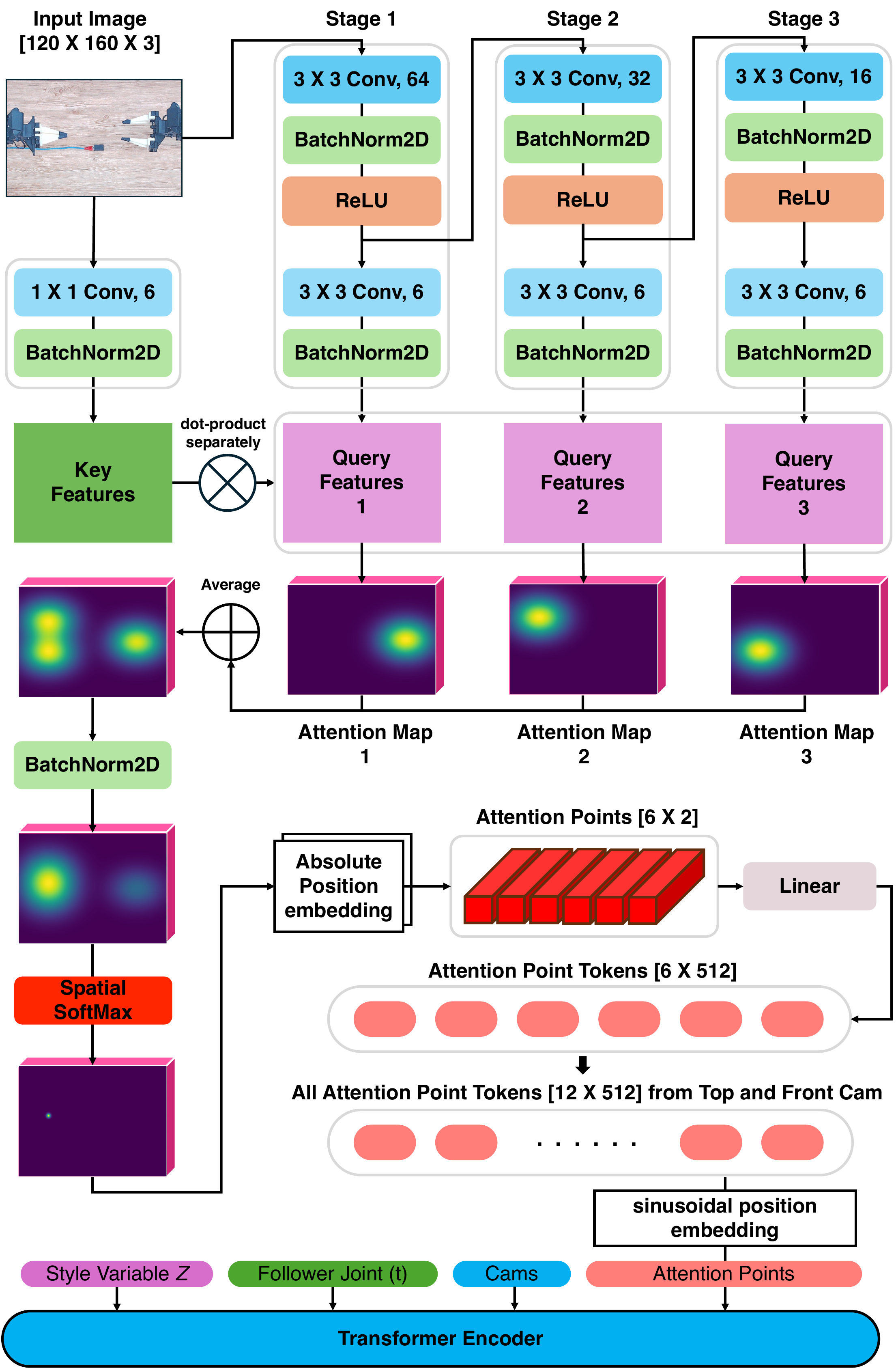}
    \caption{Detailed architecture of the multistage spatial attention module.}
    \label{fig:method_msa}
\end{figure}

\section{PROPOSED METHOD}
\subsection{Revisiting ACT}
The original Action Chunking Transformer formulates visuomotor policy learning as chunk-level sequences modeling \cite{zhao2023learning}. Instead of predicting a single action $a_t$, ACT predicts a horizon-length chunk $\mathbf{a}_{t:t+K-1}$, where $K$ is the chunk length. The action sequence reconstruction loss is $\mathcal{L}_{rec}=\| a_{t:t+K-1} - \hat{a}_{t:t+K-1} \|_1$. During execution, ACT applies temporal ensembling: $\hat{a}_t = \sum_{i=0}^{K-1} w_i \mathbf{a}_{t-i}^{(i)}$, where $\mathbf{a}{t-i}^{(i)}$ denotes the $i$-th action predicted from observation $o{t-i}$, and $w_i$ are normalized weights.  This reduces covariate shift and smooths execution. Training follows a CVAE formulation \cite{sohn2015learning}. The encoder produces latent $z$ from ground-truth action sequencess, and the decoder predicts action chunks conditioned on $(o_t, z)$. The loss is: $\mathcal{L}_{ACT}=\mathcal{L}_{rec}+\beta D_{KL}\left(q(z|a)\,\|\, p(z)\right)$. While ACT achieves strong low-latency control, it does not explicitly constrain where the model attends spatially, leaving the attention–action mapping implicit and potentially unstable under visual disturbances.

\subsection{Overall Framework: ACT with Explicit Spatial Alignment}
MSACT builds upon the original ACT framework by preserving its three core design principles: (1) Pretrained visual backbone: a ResNet encoder extracts image features as a strong visual prior. (2) Conditional VAE (CVAE) training objective: reconstruction loss for action chunks combined with KL regularization $D_{KL}$. (3) Action chunking with temporal ensembling during inference to stabilize closed-loop execution at low latency.

Unlike ACT, which directly feeds global visual features into the Transformer, we introduce a multistage spatial attention (MSA) module that extracts task-relevant 2D attention points as a structured local spatial modality. These attention points are inputted into the Transformer encoder alongside image features and robot states. Furthermore, the model jointly predicts: (1) a future K-step action sequences, (2) a future K-step Cam Top attention point sequences, (3) a future K-step Cam Front attention point sequences. Since no keypoint annotation is provided, we enforce a temporally consistency loss between predicted and re-extracted attention point sequences from ground-truth future frames. This suppresses attention drift and stabilizes the vision-to-action mapping under limited demonstrations. The overall pipeline is illustrated in Fig.\ref{fig:method_overview}.

\subsection{Multistage Spatial Attention Module}
To improve performance in  fine manipulation tasks under visual disturbances, we introduce a multistage spatial attention (MSA) module (Fig.~\ref{fig:method_msa}). At each timestep $t$, the module extracts $N=6$ task-relevant attention points from Cam Top and Cam Front images separately. To balance computational efficiency and spatial fidelity, images are downsampled from $480\times640\times3$ to $120\times160\times3$ for attention extraction. Two independent MSAs are used for the two views.

\subsubsection{Multistage Feature Extraction}
Previous spatial attention methods \cite{levine2016end}\cite{ichiwara2021spatial}\cite{lenz2015deep}\cite{ichiwara2022contact} generate attention maps from only the final feature layer, forming a single-scale representation. In contrast, the MSA adopts a three-stage CNN hierarchy, enabling multiscale perception. We define: (1) Key feature: a linear projection of the RGB input preserving spatial layout. (2) Query features: output from each CNN stage containing progressively abstract task representations. Each stage consists of 3×3 Conv + BatchNorm2D + ReLU for features extraction. All Query and Key features are linear projected to the same channel dimension $N$ using 1×1 convolutions. 

\subsubsection{Cross-Stage Attention}
For each stage, like\cite{vaswani2017attention} dot-product attention is computed and spatial normalization is applied to stabilize gradients:
{\footnotesize
\begin{equation}
Attention~map=\frac{Query \cdot Key}{\sqrt{H_{img} W_{img}}}
\end{equation}
}
where $H_{img}$ and $W_{img}$ denote the input image height and
width. This produces three attention maps corresponding to different receptive fields.

\subsubsection{Multistage Fusion}
The three maps are averaged and followed by BatchNorm2D to normalize response distributions:
{\footnotesize
\begin{equation}
A_{fused}=BatchNorm2D\left( \frac{1}{3}\sum_{s=1}^{3}Attention~map_s\right)
\end{equation} 
}
This design enables coarse-to-fine spatial detail while preserving absolute spatial coordinates\cite{lin2017feature}.

\subsubsection{Attention Points Generation}
To obtain attention points $P$, we apply temperature-controlled 2D softmax with absolute positional
embeddings $PE_{h,w}$, which is a differentiable soft-argmax operation\cite{levine2016end}:
{\footnotesize
\begin{equation}
    PE_{h,w} = \left( \frac{h}{H_{\mathrm{img}}-1},\; \frac{w}{W_{\mathrm{img}}-1} \right),
    \label{eq:absolute_positional_embedding}
\end{equation}
}
\vspace{-6pt}
{\footnotesize
\noindent  where $0 \le h < H_{\mathrm{img}}$ and $0 \le w < W_{\mathrm{img}}$
}

{\footnotesize
\begin{equation}
    P = Softmax2D\left( \frac{A_{fused}}{T} \right) \odot PE_{h,w}, 
    \qquad T = 0.001
    \label{eq:attention_points}
\end{equation}
}
Repeating for $N=6$ channels yields $6\times2$ spatial coordinates (x,y) per view. These coordinates are linearly projected into 512 dimensional tokens. Each coordinate corresponds to one token, yielding a total of 12 attention point tokens per timestep. The attention point tokens undergo sinusoidal position embedding before concatenation with latent $z$, follower joint tokens and camera feature tokens.

\subsection{Action and Attention Point Sequences Prediction}
Unlike ACT which predicts only actions, MSACT's decoder outputs: $
\{\hat{\mathbf{a}}_{t:t+K-1},\hat{\mathbf{p}}^{top}_{t:t+K-1}, \hat{\mathbf{p}}^{front}_{t:t+K-1}\}$. During training, weight-sharing MSA modules extract attention point sequences from ground-truth future frames: $\hat{\mathbf{p}}^{gt}_{t:t+K-1}$. Without keypoint annotations, supervision is self-supervised through temporal alignment. We define a temporally consistency L1 loss over the attention point sequences:
{\footnotesize
\begin{equation}
\mathcal{L}_{att}
=
\left\|
\hat{\mathbf{p}}_{t:t+K-1}
-
\hat{\mathbf{p}}^{gt}_{t:t+K-1}
\right\|_1
\end{equation}
}
This enforces forward alignment, and reduces drift accumulation. End-to-end learning without adding $\mathcal{L}_{att}$, MSA fails to extract stable attention points. The possible explanation is that ACT’s one-step sampling strategy may limit the extraction of temporally consistent attention features. The total loss extends ACT: 
{\footnotesize
\begin{equation}
\mathcal{L}
=
\mathcal{L}_{rec}
+
\beta D_{KL}
+
\mathcal{L}_{att}
\end{equation}
}

\begin{figure}
    \centering
    \includegraphics[width=1\linewidth]{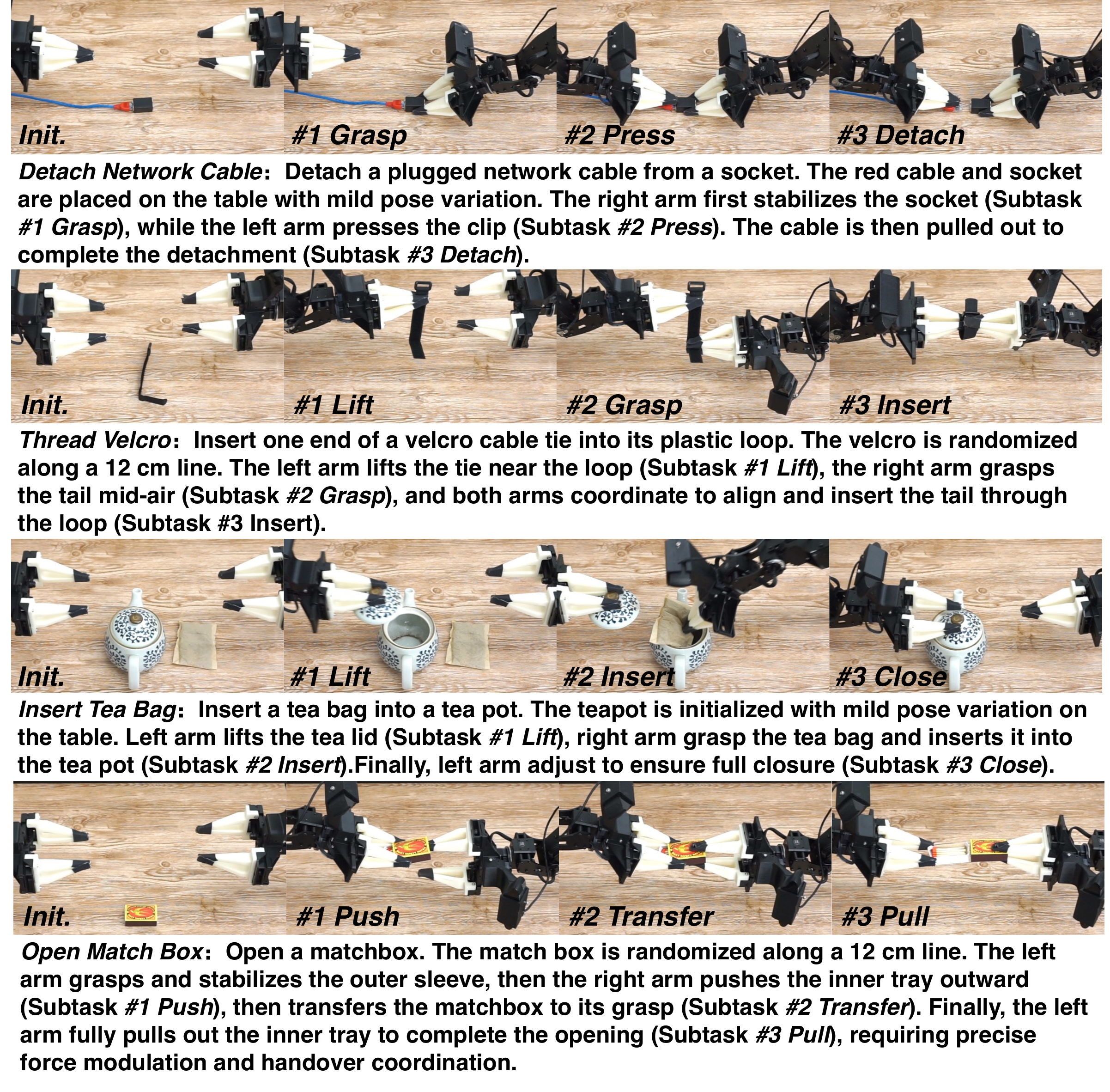}
    \caption{Real-World Task Setting. For each of the 4 real-world tasks, we illustrate the initializations and the subtasks.}
    \label{fig:tasks_overview}
\end{figure}

\begin{table}[t]
\centering
\caption{Training Settings Across Methods}
\label{tab:traing_settings}
\resizebox{0.48\textwidth}{!}{
\small
\renewcommand{\arraystretch}{1.1}
\setlength{\tabcolsep}{3pt}

\begin{tabular}{lccccc}
\toprule
\textbf{Method} &
\makecell{\textbf{Training}\\\textbf{steps}} &
\makecell{\textbf{Batch}\\\textbf{size}} &
\textbf{Optimizer} &
\makecell{\textbf{Learning}\\\textbf{rate}} &
\makecell{\textbf{Weight}\\\textbf{decay}} \\
\midrule
MSACT (SA)       & 100K & 8  & AdamW & 1e-5          & 1e-4 \\
\textbf{MSACT (Ours)}  & 100K & 8  & AdamW & 1e-5          & 1e-4 \\
ACT                & 100K & 8  & AdamW & 1e-5          & 1e-4 \\
Diffusion Policy  & 200K & 32 & Adam  & \phantom{0}1e-4$^{*}$    & 1e-6 \\
SmolVLA (0.45B)    & 30K  & 32 & AdamW & \phantom{0}\phantom{0}1e-4$^{**}$   & \phantom{0}1e-10 \\
$\pi0.5$ (3.6B)   & 30K  & 32 & AdamW & \phantom{0}2.5e-5$^{**}$ & 1e-2 \\
\bottomrule
\end{tabular}
}
\vspace{2pt}
\scriptsize
\textsuperscript{*} Cosine decay (500 warmup steps). \\
\textsuperscript{**} Cosine decay (1K warmup, 30K decay, final LR 2.5e-6).
\end{table}

\section{EXPERIMENTS}
All experiments are conducted on the ALOHA bimanual platform \cite{zhao2023learning} adopted by ACT. The system consists of two 7-DoF low-cost arms with parallel grippers, synchronized at 50 Hz control frequency. RGB images are captured from multiple cameras at $480\times640$ resolution. We evaluate on simulated and real-world bimanual tasks to assess fine-grained control, spatial alignment, and robustness under visual variation. For each real-world task, we collect 50 human teleoperated demonstrations, consistent with ACT’s limited data setting. Each demonstration is recorded for 20 seconds at randomized initial states along a 12 cm line, resulting in 1,000 time steps. The images include views from the top, front, left arm, and right arm.

We adopt the two simulated tasks introduced in ACT \cite{zhao2023learning} (Cube Transfer and Bimanual Insertion) and conduct ablation study to validate the independent effectiveness of the proposed MSA module. For simulated tasks, the model receives only top-view RGB image as visual input. We follow the ACT evaluation methodology, conducting training and testing through both scripted and human data. Each test is run 50 times using three random seeds to obtain the average success rate. For ablation study, we replace the MSA in MSACT with spatial attention (SA) module in related work \cite{ichiwara2021spatial}\cite{lenz2015deep}\cite{ichiwara2022contact} as the ablation model. 

We compare our method against representative state-of-the-art methods (ACT, Diffusion Policy, SmolVLA (0.45B) and $\pi0.5$ (3.6B)) across four real-world tasks (Fig.~\ref{fig:tasks_overview}): (1) Detach Network Cable: the right arm stabilizes the socket, the left arm presses the clip, and the cable is pulled out. (2) Thread Velcro: The left arm lifts the cable tie, the right arm grasps the tail, and both arms coordinate insertion. (3) Insert Tea Bag: The left arm lifts the lid, the right arm inserts the tea bag, and the lid is closed. (4) Open Match Box: The left arm stabilizes the outer sleeve, the right arm pushes the inner tray, and both arms coordinate to complete opening. These tasks involve: precise force modulation, object stabilization, mid-air coordination and handover behaviors. Each real-world task is evaluated independently over 100 trials with randomized initial states. Success in the final subtask is regarded as success for the entire task.

All methods are deployed on an NVIDIA RTX 4090 GPU using the LeRobot framework \cite{cadene2024lerobot}. All methods are trained or fine-tune ($\pi0$: action expert and projections only) on the same dataset. All baseline methods are trained using their officially recommended configurations and hyperparameters. The validation performance curves confirm that all methods reach plateau. Table \ref{tab:traing_settings} shows the training settings. All methods share the same chunk size or planning horizon, identical observation history (1 step) and robot control frequency (50Hz). To satisfy the real-time constraint, standard inference-time optimizations are applied consistently (DP: 10 DDIM denoising steps, $\pi0.5$: cache enabled). 

\section{Results and Discussion}

\begin{figure}
    \centering
    \includegraphics[width=1\linewidth]{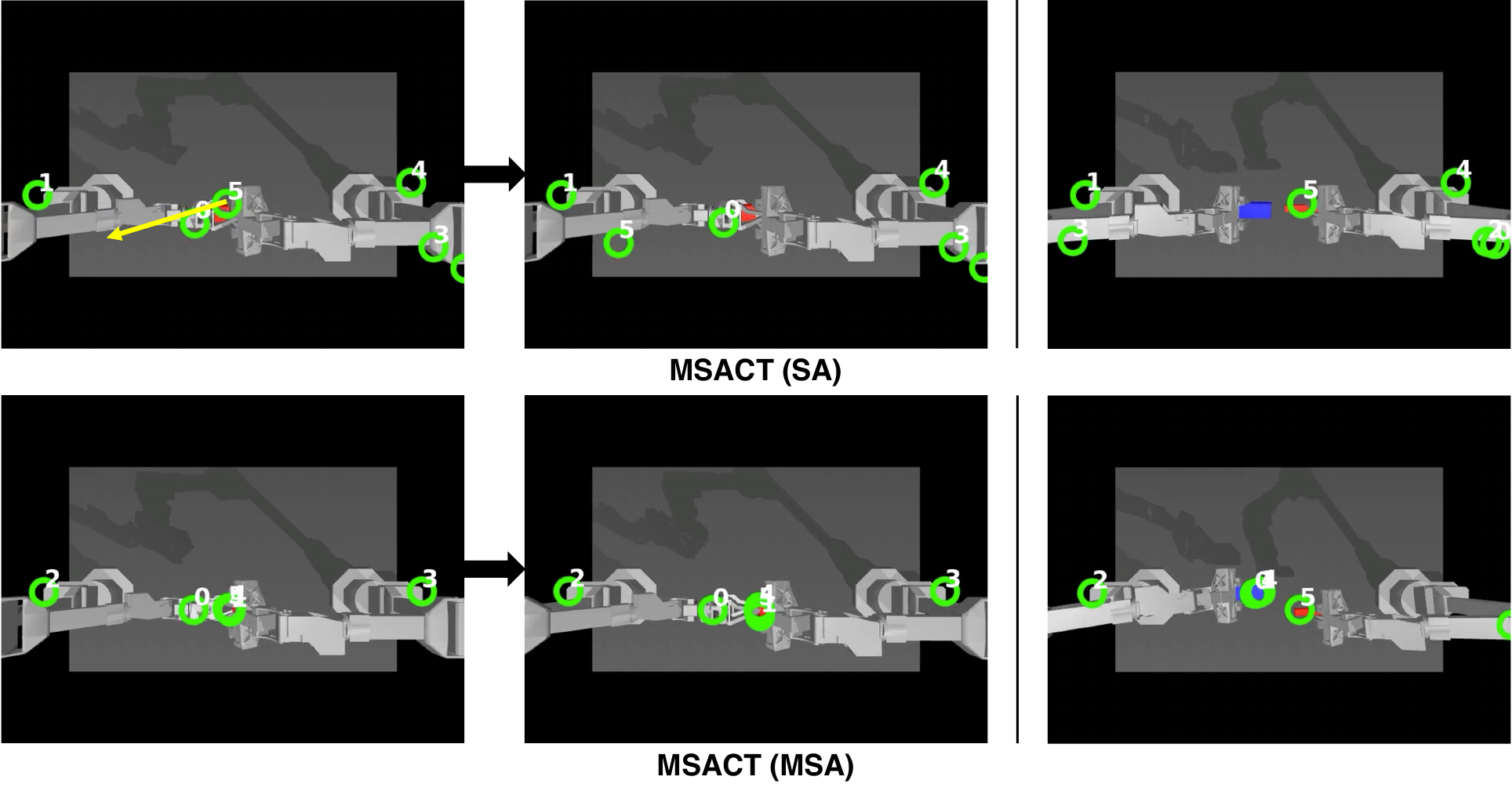}
    \caption{Visualization of attention points on the proposed and ablation model for simulated tasks. Green circles are predicted points. The yellow arrow indicates the offset position of the attention point at the next time step.}
    \label{fig:sim_attn}
\end{figure}

\begin{table}[t]
\centering
\caption{Success Rate for Simulated Tasks}
\label{tab:sim_results}
\footnotesize
\setlength{\tabcolsep}{3.5pt}
\renewcommand{\arraystretch}{1.05}

\begin{tabular}{lcccccc}
\toprule
& \multicolumn{3}{c}{\textbf{Cube Transfer (sim)}} 
& \multicolumn{3}{c}{\textbf{Bimanual Insertion (sim)}} \\
\cmidrule(lr){2-4} \cmidrule(lr){5-7}
& Touch & Lift & Transfer 
& Grasp & Contact & Insert \\
\midrule

ACT 
& \phantom{0}97$\mid$82
& \phantom{0}90$\mid$60
& \phantom{0}86$\mid$50
& \phantom{0}93$\mid$76
& \textbf{\phantom{0}90}$\mid$66
& \phantom{0}32$\mid$20 \\

MSACT (SA)
& \textbf{100}$\mid$88
& \phantom{0}99$\mid$73 
& \phantom{0}93$\mid$51
& \textbf{\phantom{0}98}$\mid$77
& \phantom{0}84$\mid$64 
& \phantom{0}44$\mid$17 \\

\textbf{MSACT (MSA)} 
& \textbf{100$\mid$95} 
& \textbf{100$\mid$76} 
& \textbf{100$\mid$76} 
& \textbf{\phantom{0}98$\mid$84} 
& \textbf{\phantom{0}90$\mid$74} 
& \textbf{\phantom{0}49$\mid$26} \\

\bottomrule
\end{tabular}
\end{table}

\subsection{Results on Simulated Tasks}
Table \ref{tab:sim_results} shows success rates on simulated tasks under scripted and human data ($Left$$\mid$$Right$). Our method improves overall performance across tasks. In Cube Transfer, improvements are observed in Touch, Lift, and Transfer stages, with near-saturated performance under scripted data and clear gains under human demonstrations. In Bimanual Insertion, performance is also improved in Grasp and Insert, though the margin is smaller than in Cube Transfer.

Fig.~\ref{fig:sim_attn} visualizes predicted attention points. Compared with SA, which shows less stable and occasionally diffused localization, the MSA module focuses on task-relevant objects and grippers while maintaining more consistent localization across stages. In Cube Transfer, attention shifts coherently from contact to handover and reception. In Bimanual Insertion, attention tracks both peg and socket during alignment.

The relatively smaller gain in Bimanual Insertion is attributed to limited geometric observability. Simulated tasks use only a top-view image, which lacks sufficient depth cues for mid-air alignment. Although spatial localization is improved, the absence of complementary viewpoints constrains insertion accuracy. This motivates our real-world setup, where Top and Front views jointly provide more reliable distance information to support aerial alignment.

\begin{table}[t]
\centering
\caption{Success Rate for Real-world Tasks}
\footnotesize
\setlength{\tabcolsep}{4pt}
\renewcommand{\arraystretch}{1.1}

\begin{tabular}{lccc ccc}
\toprule
 & \multicolumn{3}{c}{\textbf{Detach Network Cable}}
 & \multicolumn{3}{c}{\textbf{Thread Velcro}} \\
\cmidrule(lr){2-4}\cmidrule(lr){5-7}
\textbf{Model}
& Grasp & Press & Detach
& Lift & Grasp & Insert \\
\midrule
DP        & 70 & 6  & 0  & 48 & 0  & 0  \\
ACT       & \textbf{100} & 46 & 26 & 80 & 68 & 8  \\
SmolVLA (0.45B)  & \textbf{100} & 50 & 2  & 58 & 51 & 3  \\
$\pi0.5$ (3.6B) & \textbf{100} & 51 & 26 & 83 & 56 & 5  \\
\textbf{MSACT (Ours)}
          & \textbf{100} & \textbf{88} & \textbf{72}
          & \textbf{87} & \textbf{82} & \textbf{21} \\
\bottomrule
\end{tabular}

\vspace{6pt}

\begin{tabular}{lccc ccc}
\toprule
 & \multicolumn{3}{c}{\textbf{Insert Tea Bag}}
 & \multicolumn{3}{c}{\textbf{Open Match Box}} \\
\cmidrule(lr){2-4}\cmidrule(lr){5-7}
\textbf{Model}
& Lift & Insert & Close
& Push & Transfer & Pull \\
\midrule
DP        & 0  & 0  & 0  & 0  & 0 & 0  \\
ACT       & 65 & 46 & 21 & 76 & 47 & 38 \\
SmolVLA (0.45B)  & 74 & 73 & \textbf{56} & 12 & 0  & 0  \\
$\pi0.5$ (3.6B) & 89 & 62 & 15 & 42 & 22 & 6  \\
\textbf{MSACT (Ours)}
          & \textbf{100} & \textbf{95} & \textbf{56}
          & \textbf{94} & \textbf{85} & \textbf{63} \\
\bottomrule
\end{tabular}

\label{tab:real_subtasks}
\end{table}

\begin{figure*}
    \centering
    \includegraphics[width=0.9\linewidth]{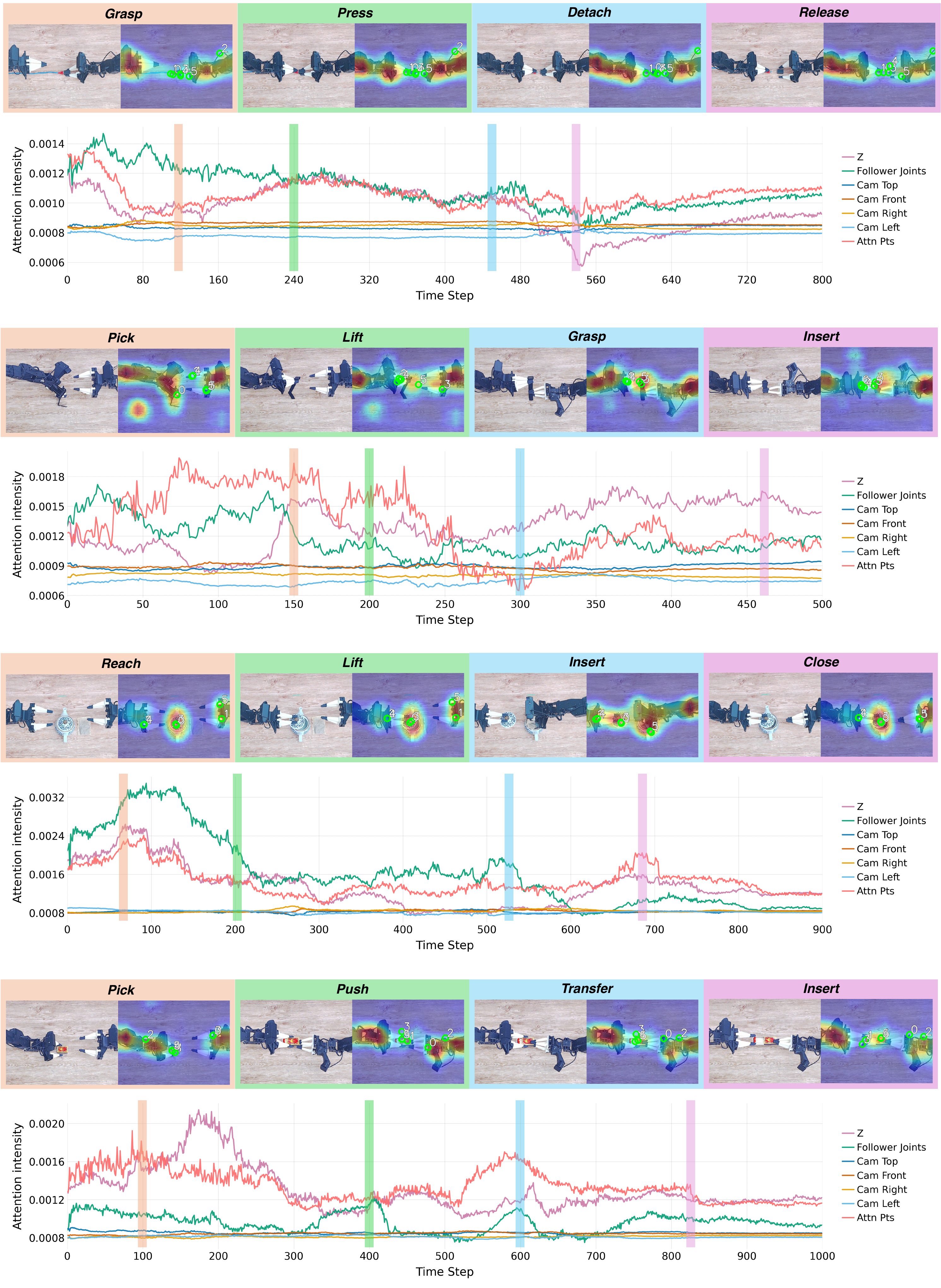}
    \caption{Visualization of attention (attention points and averaged image features extracted by ResNet) on the proposed model for 4 real-world tasks (Detach Network Cable, Thread Velcro, Insert Tea Bag and Open Match Box), and attention intensity of modalities. Green circles are predicted points. For average image features, the color goes from blue to red, the higher its value.}
    \label{fig:attn_results}
\end{figure*}

\subsection{Results on Real-world Tasks}

Table \ref{tab:real_subtasks} shows success rate of real-world tasks. Across all four tasks, our method achieves the highest overall success rates compared with ACT, Diffusion Policy, SmolVLA, and $\pi0.5$. In Detach Network Cable, improvements are most evident in the Press and Detach stages, where coordinated force application and contact timing are critical. The method achieves higher completion counts than ACT and $\pi0.5$. In Thread Velcro, the largest gain is observed in the Insert stage, which requires precise mid-air alignment and bimanual coordination. Baseline methods often fail due to misalignment. In Insert Tea Bag, performance is near-saturated in the Lift and Insert stages, with stable execution during closing. Compared with ACT and SmolVLA, the proposed method exhibits more consistent action sequencing under pose variation. In Open Match Box, improvements appear in the Transfer and Pull stages, suggesting better handover stability and spatial alignment during object exchange. Overall, the method improves both stage-wise and full-task completion across all real-world tasks, indicating increased robustness under limited data.

Fig.~\ref{fig:attn_results} visualizes predicted attention points and averaged image features extracted by ResNet. The attention points consistently concentrates on task-relevant objects and grippers, remaining stable throughout task progression rather than drifting toward background regions. Stage-wise visualizations further show that attention points tracks object contacts and transitions while maintaining coherent spatial localization.

\begin{table}[t]
\centering
\caption{Inference Latency and Overall Success Rate}
\footnotesize
\setlength{\tabcolsep}{4pt}
\renewcommand{\arraystretch}{1.1}

\begin{tabular}{lccc}
\toprule
\textbf{Model} & \textbf{Latency (ms)} & \textbf{Success Rate (\%)} &  \textbf{[99\% CI]}\\
\midrule

DP
& 158.1$\pm$14.9 
& \phantom{0}0.00
& [\phantom{0}0.00-1.63] \\

ACT 
& 45.34$\pm$1.03 
& 23.25
& [18.27-29.10] \\

SmolVLA
& 91.23$\pm$1.34 
& 15.25
& [11.19-20.44] \\

$\pi0.5$ 
& 112.1$\pm$0.40 
& 13.00 
& [\phantom{0}9.27-17.94] \\

\textbf{MSACT (Ours)} 
& 45.40$\pm$5.00 
& 53.00
& [46.58-59.33] \\
\bottomrule
\end{tabular}
\label{tab:latency_real}
\end{table}

\subsection{Inference Latency and Real-Time Performance}
Table \ref{tab:latency_real} shows the inference latency for all methods. Latency is defined as the end-to-end elapsed time from the availability of the latest camera frames (four 480×640 RGB images) to the output of the policy action, including image preprocessing, GPU transfer, model inference, and post-processing. Inference latency is reported as mean ± standard deviation over 400 forward passes (4 real-world tasks × 100 trials). The overall success rate is computed by aggregating all 400 trials and treating them as independent Bernoulli outcomes. We report the overall task success rate with 99\% confidence intervals computed using the Wilson score method. MSACT exhibits inference latency comparable to ACT, operating at 45.40 ± 5.00 ms per inference compared to 45.34 ± 1.03 ms for ACT. The additional computation introduced by MSA consists of lightweight 1×1 convolutions and spatial softmax operations. These operations are small relative to the ResNet backbone and Transformer forward pass. Despite this similar computational cost, MSACT achieves a higher real-world success rate (53.00\%) compared with ACT (23.25\%). A two-sided Fisher’s exact test on the aggregated outcomes indicates that the improvement of MSACT over ACT is statistically significant ($p < 0.001$).

\subsection{Performance under Visual Disturbance}
\begin{figure}[!t]
    \centering
    \includegraphics[width=0.85\linewidth]{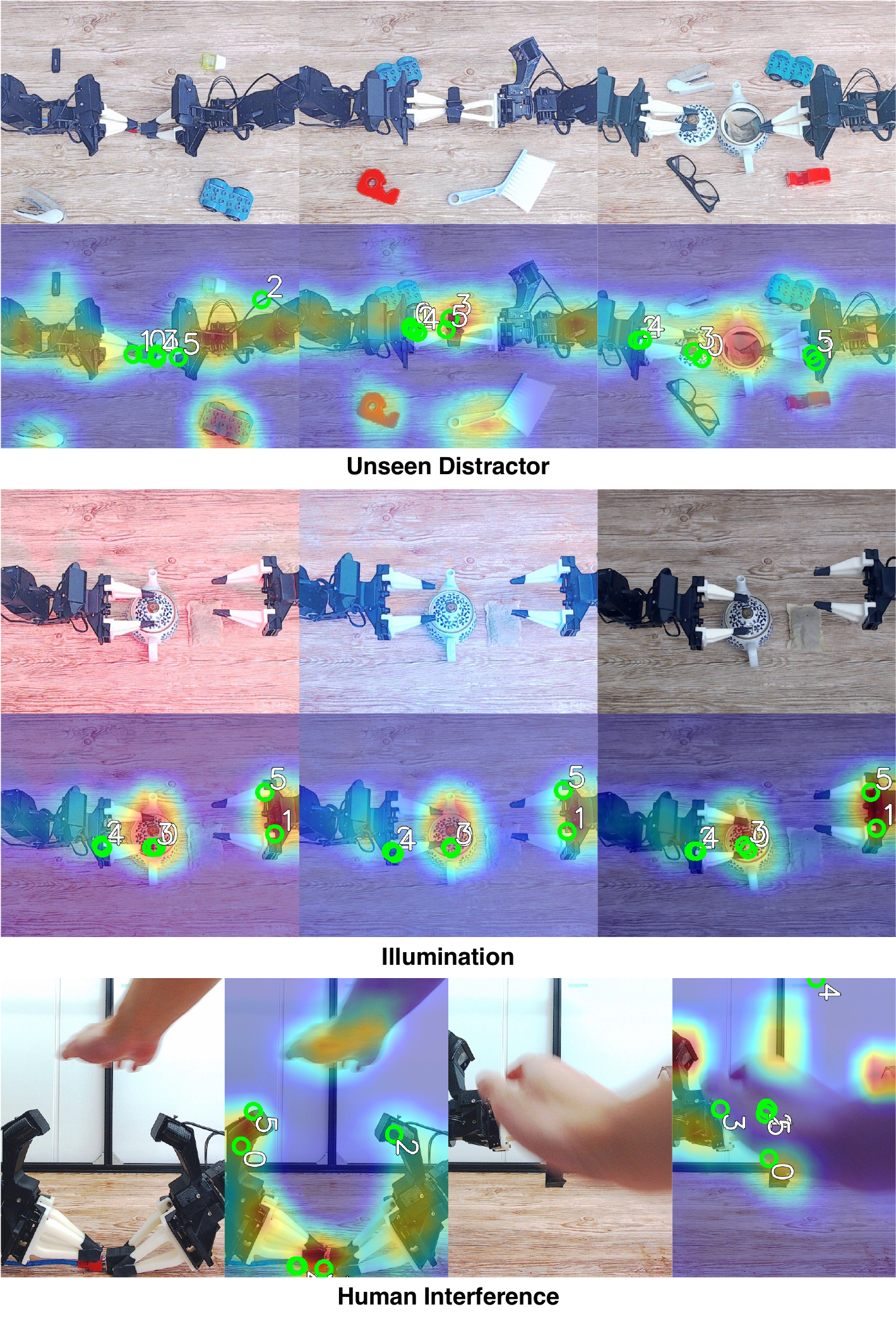}
    \caption{Proposed method's attention under different visual disturbances.}
    \label{fig:robustness}
\end{figure}

We evaluate the performance of the proposed model to visual disturbances under three conditions (Fig.~\ref{fig:robustness}). (1) unseen distractors randomly placed in the workspace, (2) illumination changes (color tint and low light), and (3) dynamic human hand interference. Although the ResNet backbone shows slight attention bias toward salient or moving distractors, the predicted task-relevant attention points remain stable and focused on the manipulation target, enabling reliable execution under object-level, appearance, and dynamic disturbances.
Using the Insert Tea Bag task as an example, each condition is tested 30 times. Our model maintains success rates of 53\% under illumination changes and 50\% under human interference, while ACT drops to 17\% in both case. With unseen distractor, our success rate decreases to 43\%, compared to ACT’s 10\%.

\subsection{Attention-point Modality Analysis}
To further analyze the influence of the attention-point modality on action prediction, we visualize the cross-attention values in the Transformer decoder's last layer. Since different modalities contain different numbers of tokens (style variable latent $z$:1, follower joints:1, camera views:$4\times300$, attention points:12), directly summing attention weights would bias comparison toward longer token sequences. To obtain a more balanced estimate of modality contribution, we compute a normalized attention intensity for each modality: $I_m = \frac{\sum_{i \in \mathcal{T}_m} \alpha_i}{|\mathcal{T}_m|}$, where $\mathcal{T}_m$ denotes the token set of modality $m$, $\alpha_i$ is the cross-attention weight, and $|\mathcal{T}_m|$ is its token length.

The resulting attention intensity curves (bottom plots in Fig.~\ref{fig:attn_results}) show that the attention-point modality consistently exhibits elevated attention during key interaction phases of manipulation. In Detach Network Cable, attention gradually shifts from follower-joint signals toward attention points during contact-rich stages (Press and Detach). In Thread Velcro, attention points show higher intensity during early object localization (Pick, Lift), while the style latent $z$ becomes more prominent during mid-air coordination and insertion. In Insert Tea Bag, attention-point intensity increases during reaching and closing stages, which involve precise spatial alignment. In Open Match Box, attention-point intensity peaks during pick and transfer stages, whereas follower-joint signals become more prominent during insertion, suggesting a transition from coarse positioning to fine motion control.

Across tasks, the attention-point modality exhibits stage-aligned intensity variations that are consistent with the temporal structure of manipulation. In contrast, camera image modalities show comparatively lower and less stage-specific attention. This suggests that the explicitly local spatial modality may provide structured geometric information that complements visual inputs during action decoding. While attention weights do not directly imply causal importance, the consistent alignment between attention-point intensity and task phases indicates that the MSA module produces spatial representations that are effectively utilized by the policy network. These results are consistent with the role of multistage spatial alignment in facilitating stable and low-latency fine manipulation by providing temporally coherent spatial references across interaction stages.

\section{CONCLUSIONS AND FUTURE WORK}
We presented MSACT, which introduces multistage spatial alignment for stable low-latency fine manipulation. Experiments in simulation and real-world bimanual tasks show consistent performance gains, particularly in contact-rich interactions and mid-air coordination. These results suggest that structured spatial alignment across stages can facilitate precise manipulation under latency constraints. Future work will explore integrating this approach into vision–language–action models, including language-conditioned modulation of attention points to support more complex multi-task behaviors.

\addtolength{\textheight}{-12cm}   



\section*{ACKNOWLEDGMENT}
This work was supported by the JST Moonshot Research and Development Project (JPMJMS2031) and the Research Institute of Science and Engineering, Waseda University. We would like to express our gratitude for this support.

\bibliographystyle{IEEEtran} 
\bibliography{refs}
\end{document}